\documentclass{article}
\usepackage{spconf,amsmath,graphicx}
\usepackage{amsfonts}
\usepackage{url}
\usepackage{multirow,booktabs}
\usepackage[table,xcdraw]{xcolor}

\title{Deja-vu: Double Feature Presentation and Iterated Loss \\in Deep Transformer Networks}
\name{\parbox{\linewidth}{\centering Andros Tjandra$^{1*}$, Chunxi Liu$^{2}$,  Frank Zhang$^{2}$, Xiaohui Zhang$^{2}$, Yongqiang Wang$^{2}$,\\Gabriel Synnaeve$^{2}$, Satoshi Nakamura$^1$, Geoffrey Zweig$^{2}$\thanks{* This work was done while the first author was a research intern at Facebook.}}}

\address{$^1$Nara Institute of Science and Technology, Japan \\ $^2$Facebook AI, USA \\
\small{\texttt{\{andros.tjandra.ai6,s-nakamura\}@is.naist.jp,}}\\ \small{\texttt{\{chunxiliu,frankz,xiaohuizhang,yqw,gab,gzweig\}@fb.com}}
}
\begin{document}
\ninept
\maketitle
\begin{abstract}
Deep acoustic models typically receive features in the first layer of the network, and process increasingly abstract representations in the subsequent layers. Here, we propose to feed the input features at multiple depths in the acoustic model. As our motivation is to allow acoustic models to re-examine their input features in light of partial hypotheses we introduce intermediate model heads and loss function. We study this architecture in the context of deep Transformer networks, and we use an attention mechanism over both the previous layer activations and the input features. To train this model's intermediate output hypothesis, we apply the objective function at each layer right before feature re-use. We find that the use of such iterated loss significantly improves performance by itself, as well as enabling input feature re-use. We present results on both Librispeech, and a large scale video dataset, with relative improvements of 10 - 20\% for Librispeech and 3.2 - 13\% for videos.
\end{abstract}
\section{Introduction}
\label{sec:intro}

In this paper, we propose the processing of features not only in the input layer of a deep network, but in the intermediate layers as well. We are motivated by a desire to enable a neural network acoustic model to adaptively process the features depending on partial hypotheses and noise conditions. Many previous methods for adaptation have operated by linearly transforming either input features or intermediate layers in a two pass process where the transform is learned to maximize the likelihood of some adaptation data 
\cite{Yu2013adaptation,delcroix2015context,yao2012adaptation}.
Other methods have involved characterizing the input via factor analysis or i-vectors \cite{li2014factorized,saon2013speaker}. Here, we suggest an alternative approach in which adaptation can be achieved by re-presenting the feature stream at an intermediate layer of the network that is constructed to be correlated with the ultimate graphemic or phonetic output of the system.

We present this work in the context of Transformer networks \cite{vaswani2017attention}. Transformers have become a popular deep learning architecture for modeling sequential datasets, showing improvements in many tasks such as machine translation \cite{vaswani2017attention} and language modeling \cite{dai2019transformer}. In the speech recognition field, Transformers have been proposed to replace recurrent neural network (RNN) architectures such as long short-term memory (LSTMs) and gated recurrent units (GRUs) \cite{mohamed2019transformers}. A recent survey of Transformers in many speech related applications may be found in \cite{karita2019comparative}. Compared to RNNs, Transformers have several advantages, specifically an ability to aggregate information across all the time-steps by using a self-attention mechanism. Unlike RNNs, the hidden representations do not need to be computed sequentially across time, thus enabling significant efficiency improvements via parallelization. 

In the context of Transformer module, secondary feature analysis is enabled through an additional mid-network transformer module that has access both to previous-layer activations and the raw features. To implement this model, we apply the objective function several times at the intermediate layers, to encourage the development of phonetically relevant hypotheses. Interestingly, we find that the iterated use of an auxiliary loss in the intermediate layers significantly improves performance by itself, as well as enabling the secondary feature analysis. 

This paper makes two main contributions:
\begin{enumerate}
\item We present improvements in the basic training process of deep transformer networks, specifically the iterated use of connectionist temporal classification (CTC) or cross-entropy (CE) in intermediate layers, and
\item We show that an intermediate-layer attention model with access to both previous-layer activations and raw feature inputs can significantly improve performance.
\end{enumerate}

We evaluate our proposed model on Librispeech and a large-scale video dataset. From our experimental results, we observe 10-20\% relative improvement on Librispeech and 3.2-11\% on the video dataset.

\section{Transformer Modules}
A transformer network \cite{vaswani2017attention} is a powerful approach to learning and modeling sequential data. A transformer network is itself constructed with a series of transformer modules that each perform some processing. Each module has a self-attention mechanism and several feed-forward layers, enabling easy parallelization over time-steps compared to recurrent models such as RNNs or LSTMs \cite{hochreiter1997long}. We use the architecture defined in \cite{vaswani2017attention}, and provide only a brief summary below.

Assume we have an input sequence that is of length $S$: $X = [x_1,...,x_S]$. Each $x_i$ is itself a vector of activations. A transformer layer encodes $X$ into a corresponding output representation $Z = [z_1,...,z_S]$ as described below.

\begin{figure}[h]
\caption{A Transformer Module.}\label{fig:transformer_block}
\centering
\includegraphics[width=0.25\columnwidth]{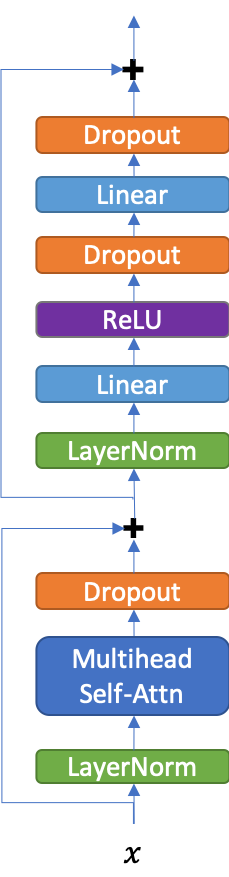}
\vspace{-0.3cm}
\end{figure}

Transformers are built around the notion of a self-attention mechanism that is used to extract the relevant information for each time-step $s$ from all time-steps $[1..S]$ in the preceding layer. Self attention is defined in terms of a Query, Key, Value triplet $\{{Q}, {K}, {V}\} \in \mathbb{R}^{S \times d_k}$. In self-attention, the queries, keys and values are the columns of the input itself, $[x_1,...,x_S]$. The output activations are computed as:
\begin{align}
    \text{Attn}(Q,K,V) = \text{softmax}\left(\frac{Q\, K^{T}}{\sqrt{d_k}}\right)V.
\end{align}

Transformer modules deploy a multi-headed version of self-attention. As described in \cite{vaswani2017attention}, this is done by linearly projecting the queries, keys and values $P$ times with different, learned
linear projections. Self-attention is then applied to each of these projected versions of Queries, Keys and Values. These are concatenated and once again projected, resulting in the final values. We refer to the input projection matrices as  $W_p^{Q}, W_p^{K}, W_p^{V}$, and to the output projection as $W_O$. Multihead attention is implemented as
\begin{align}
    \text{MultiAttn}(Q, K, V) &= \text{concat}(\bar{V_1}, ..,\bar{V_P}) \, W_O \\
    \text{where} \,\, \forall p \in \{1..P\}\, , \bar{V_p} &= \text{Attn}(Q  W_p^{Q}, K W_p^{K}, V W_p^{V}).
\end{align} Here, $ W_p^Q, W_p^K, W_p^V \in \mathbb{R}^{d_{k} \times d_m}$, $d_m = d_{k} / P$, and $W_O \in \mathbb{R}^{Pd_m \times d_k}$.

After self-attention, a transformer module applies a series of linear layer, RELU, layer-norm and dropout operations, as well as the application of residual connections. The full sequence of processing is illustrated in Figure~\ref{fig:transformer_block}.

\section{Iterated Feature Presentation} \label{sec:proposed}

\begin{figure}[h]
\caption{A 24 layer transformer with one auxiliary loss and feature re-presentation in the 12-th layer. $Z_{0}$ represents the input features. Orange boxes represent an additional MLP network and softmax. Green boxes represent linear projections and layer-norm.}\label{fig:proposed_arch}
\centering
\includegraphics[width=0.7\columnwidth]{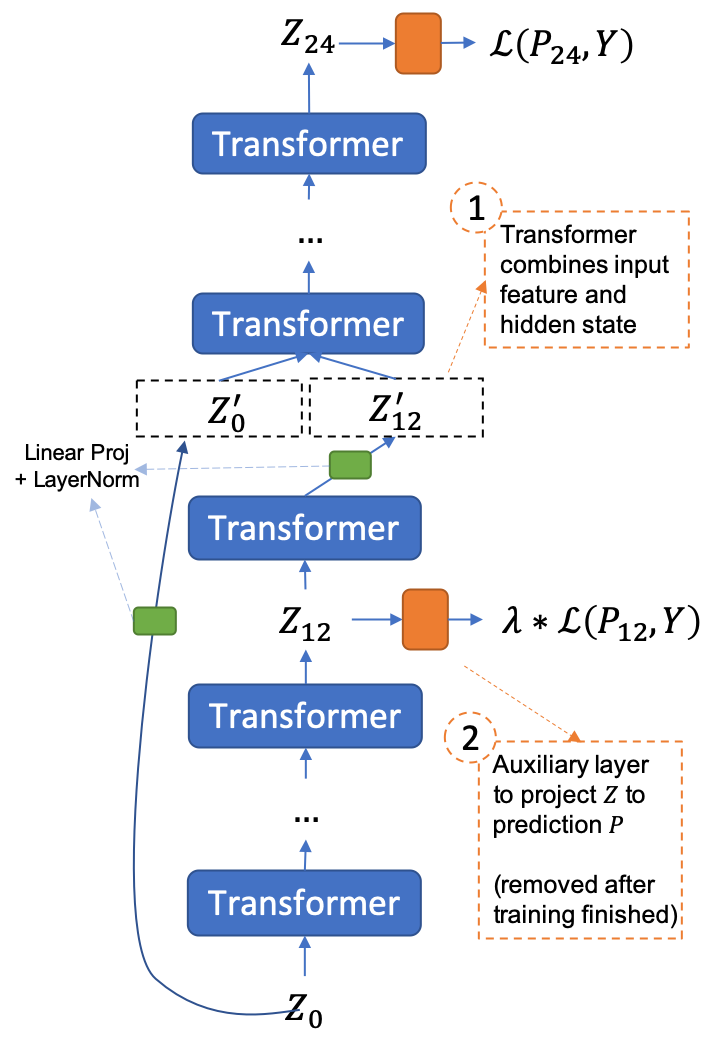}
\vspace{-0.15cm}

\end{figure}

In this section, we present our proposal for allowing the network to (re)-consider the input features in the light of intermediate processing. We do this by again deploying a self-attention mechanism to combine the information present in the original features with the information available in the activations of an intermediate layer. As described earlier, we calculate the output posteriors and auxiliary loss at the intermediate layer as well. The overall architecture is illustrated in Figure \ref{fig:proposed_arch}. Here, we have used a $24$ layer network, with feature re-presentation after the $12$th layer.

In the following subsections, we provide detail on the feature re-presentation mechanism, and iterated loss calculation.

\subsection{Feature Re-Presentation}

We process the features in the intermediate layer by concatenating a projection of the original features with a projection of previous hidden layer activations, and then applying self-attention.

First, we project both the input and intermediate layer features $(Z_0 \in \mathbb{R}^{S \times d_0}, Z_{k} \in \mathbb{R}^{S \times d_{k}} )$, apply layer normalization and concatenate with position encoding: 
\begin{align}
    Z_0^{'} = \texttt{cat}([\texttt{LayerNorm}(Z_0 W_1), E], \texttt{dim}=1) \nonumber\\
    Z_{k}^{'} = \texttt{cat}([\texttt{LayerNorm}({Z_{k} W_2}), E], \texttt{dim}=1) \nonumber
\end{align}  where $d_0$ is the input feature dimension, $d_k$ is the Transformer output dimension, $\texttt{dim}=1$ denotes concatenation on the feature axis, $W_1 \in \mathbb{R}^{d_0 \times d_c}, W_2 \in \mathbb{R}^{d_{k} \times d_c}$ and $E \in \mathbb{R}^{S \times d_{e}}$ is a sinusoidal position encoding \cite{vaswani2017attention}.

After we project both information to the same dimension, we merge them by using time-axis concatenation:
    \begin{align}
    O &= \texttt{cat}([Z_{0}^{'}, Z_{k}^{'}], \texttt{dim}=0) \in \mathbb{R}^{2S \times (d_c + d_e)}\nonumber
    \end{align}
Then, we extract relevant features with extra Transformer layer and followed by linear projection and ReLU:
    \begin{align}
      Z_{k+1}^{'} &=
    \begin{cases}
        \texttt{Transformer}(\texttt{Q}=Z_{0}^{'}, \texttt{K}=O, \texttt{V}=O), & \text{split A}\nonumber\\
        \texttt{Transformer}(\texttt{Q}=Z_{k}^{'}, \texttt{K}=O, \texttt{V}=O), & \text{split B}\nonumber
    \end{cases} \\
    Z_{k+1} &= \texttt{LayerNorm}(\texttt{ReLU}(Z_{k+1}^{'} W_3)) \nonumber
    \end{align}
    where $W_3 \in \mathbb{R}^{d_{k+1}^{'} \times d_{k+1}}$ is a linear projection.  All biases in the formula above are omitted for simplicity.
    
\begin{figure}[] 
\caption{Merging input features and intermediate layer activations with time axis concatenation for the Key and Value. Transformer layer finds relevant features based on the Query. \textbf{Split A} uses projected input features as the Query and \textbf{Split B} used projected intermediate layer activations as the Query.}\label{fig:concat_time}

\centering
\includegraphics[width=0.65\columnwidth]{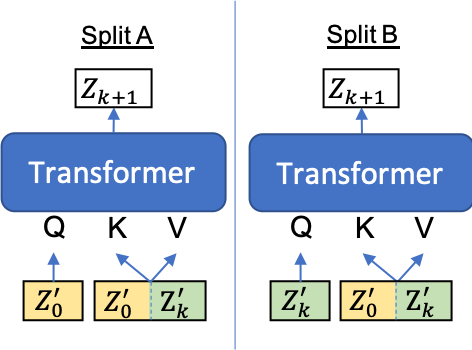}
\vspace{-0.2cm}
\end{figure}

Note that in doing time-axis concatenation, our Key and Value sequences are twice as long as the original input. In the standard self-attention where the Query is the same as the Key and Value, the output preserves the sequence length. Therefore, in order to maintain the necessary sequence length $S$, we select either the first half (split A) or the second half (split B) to represent the combined information. The difference between these two is that the use of split A uses the projected input features as the Query set, while split B uses the projected higher level activations as the Query. In initial experiments, we found that the use of high-level features (split B) as queries is preferable. We illustrate this operation on Figure~\ref{fig:concat_time}.

Another way of combining information from the features with an intermediate layer is to concatenate the two along with the feature rather than the time axis. However, in initial experiments, we found that time axis concatenation produces better results, and focus on that in the experimental results.

\subsection{Iterated Loss}
We have found it beneficial to apply the loss function at several intermediate layers of the network. Suppose there are $M$ total layers, and define a subset of these layers at which to apply the loss function:  $K = \{k_1, k_2, ..., k_L\} \subseteq \{1,..,M-1\}$. The total objective function is then defined as 
\begin{align}
    \mathcal{L} &= Loss(P_{M}, Y) + \lambda \sum_{l=1}^{L} Loss(P_{k_l}, Y) \\
    P_{k_l} &= \text{Softmax}(\text{MLP}_{l}(Z_{k_l}))
\end{align} where $Z_{k_l}$ is the $k_l$-th Transformer layer activations, $Y$ is the ground-truth transcription for CTC and context dependent states for hybrid ASR, and $Loss(P, Y)$ can be defined as CTC objective \cite{graves2006connectionist} or CE for hybrid ASR. The coefficient $\lambda$ scales the auxiliary loss and we set $\lambda = 0.3$ based on our preliminary experiments. We illustrate the auxiliary prediction and loss in Figure~\ref{fig:proposed_arch}. 

\section{Experimental results}
\subsection{Dataset}
\vspace{-0.15cm}
We evaluate our proposed module on both the Librispeech \cite{panayotov2015librispeech} dataset and a large-scale English video dataset. In the Librispeech training set, there are three splits, containing 100 and 360 hours sets of clean speech and 500 hours of other speech. We combined everything, resulting in 960 hours of training data. For the development set, there are also two splits: \texttt{dev-clean} and \texttt{dev-other}. For the test set, there is an analogous split.

The video dataset is a collection of public and anonymized English videos.
It consists of a $1000$ hour training set, a $9$ hour dev set, and a $46.1$ hour test set. The test set comprises an $8.5$ hour \texttt{curated} set of carefully selected very clean videos, a $19$ hour \texttt{clean} set and a $18.6$ hour \texttt{noisy} set \cite{le2019senones}. For the hybrid ASR experiments on video dataset, alignments were generated with a production system trained with 14k hours.

All speech features are extracted by using log Mel-filterbanks with 80 dimensions, a 25 ms window size and a 10 ms time step between two windows. Then we apply mean and variance normalization.

\subsection{Target Units}
\vspace{-0.15cm}
For CTC training, we use word-pieces as our target. During training, the reference is tokenized to 5000 sub-word units using \emph{sentencepiece}\footnote{\url{https://github.com/google/sentencepiece}} with a uni-gram language model \cite{kudo2018subword}. Neural networks are thus used to produce a posterior distribution for 5001 symbols (5000 sub-word units plus blank symbol) every frame. For decoding,
each sub-word is modeled by a HMM with two states where the last states share the same blank symbol probability; 
the best sub-word segmentation of each word is used to form a lexicon; these HMMs, lexicon are then combined with the standard $n$-gram via FST \cite{mohri2002weighted} to form a static decoding graph. Kaldi decoder\cite{povey2011kaldi} is used to produce the best hypothesis.



We further present results with hybrid ASR systems. In this, we use the same HMM topology, GMM bootstrapping and decision tree building procedure as \cite{le2019senones}. Specifically, we use context-dependent (CD) graphemes as modeling units. On top of alignments from a GMM model, we build a decision tree to cluster CD graphemes. This results in 7248 context dependent units for Librispeech, and 6560 units for the video dataset. Training then proceeds with the CE loss function. We also apply SpecAugment \cite{park2019specaug} online during training, using the LD policy without time warping. For decoding, a standard Kaldi's WFST decoder \cite{povey2011kaldi} is used.

\subsection{Deep Transformer Acoustic Model}
\vspace{-0.15cm}
All neural networks are implemented with the in-house extension of the \emph{fairseq} \cite{ott2019fairseq} toolkit. 
Our speech features are produced by processing the log Mel-spectrogram with two VGG \cite{vgg2014} layers that have the following configurations: (1) two 2-D convolutions with 32 output filters, kernel=3, stride=1, ReLU activation, and max-pooling kernel=2, (2) two 2-D convolutions with 64 output filters, kernel=3, stride=1 and max-pooling kernel=2 for CTC or max-pooling kernel=1 for hybrid. 
After the VGG layers, the total number of frames are subsampled by (i) 4x for CTC, or (ii) 2x for hybrid, thus enabling us to reduce the run-time and memory usage significantly. After VGG processing, we use 24 Transformer layers with $d_k=512$ head dimensions (8 heads, each head has 64 dimensions), 2048 feedforward hidden dimensions (total parameters 80 millions), and dropout $0.15$. For the proposed models, we utilized an auxiliary MLP with two linear layers with 256 hidden units, LeakyReLU activation and softmax (see Sec.~\ref{sec:proposed}). We set our position encoding dimensions $d_e=256$ and pre-concatenation projection $d_c=768$ for the feature re-presentation layer.
The loss function is either CTC loss or hybrid CE loss.

\subsection{Results}
\vspace{-0.15cm}
Table \ref{tab:libri-ctc} presents CTC based results for the Librispeech dataset, without data augmentation. Our baseline is a 24 layer Transformer network trained with CTC. For the proposed method, we varied the number and placement of iterated loss and the feature re-presentation. The next three results show the effect of using CTC multiple times. We see 12 and 8\% relative improvements for test-clean and test-other. Adding feature re-presentation gives a further boost, with net 20 and 18\% relative improvements over the baseline.

\addtolength{\tabcolsep}{-1pt} 
\begin{table}[]
\begin{tabular}{lccccc}
\toprule
\textbf{Model} & \textbf{Config} & \multicolumn{2}{c}{\textbf{dev}} & \multicolumn{2}{c}{\textbf{test}} \\
& & \textbf{clean} & \textbf{other} & \textbf{clean} & \textbf{other} \\
\midrule
\textbf{(CTC)}: Baseline   & VGG+24 Trf. & 4.7 & 12.7 & 5.0 & 13.1 \\ 
~~ + Iter. Loss & 12-24       & 4.1 & 11.8 & 4.5 & 12.2 \\ 
                & 8-16-24     & 4.2 & 11.9 & 4.6 & 12.3 \\ 
                & 6-12-18-24  & 4.1 & 11.7 & 4.4 & 12.0 \\
~~ + Feat. Cat. & 12-24       & 3.9 & 10.9 & 4.2 & 11.1 \\
                & 8-16-24     & 3.7 & 10.3 & 4.1 & \textbf{10.7} \\ 
                & 6-12-18-24  & 3.6 & 10.4 & \textbf{4.0} & 10.8 \\ 
\bottomrule
\end{tabular}
\caption{Librispeech CTC experimental results without any data augmentation technique and decoded with FST based on 4-gram LM. (\textbf{Notes}: Trf is Transfromers, ``+ Iter. Loss 12-24'' means adding iterative losses in the 12-th and 24-th layer, ``+ Feat. Cat. 12-24'' means adding feature concatenation in the 12-th layer.)}
\label{tab:libri-ctc}
\vspace{-0.3cm}
\end{table}

Table \ref{tab:libri-spaug} shows results for Librispeech with SpecAugment. We test both CTC and CE/hybrid systems. There are consistent gains first from iterated loss, and then from multiple feature presentation. We also run additional CTC experiments with 36 layers Transformer (total parameters 120 millions). The baseline with 36 layers has the same performance with 24 layers, but by adding the proposed methods, the 36 layer performance improved to give the best results. This shows that our proposed methods can improve even very deep models. 

\begin{table}
\begin{center}
\begin{tabular}{lccc}
\toprule
 \textbf{Model} & \textbf{LM} &  \textbf{test-clean} & \textbf{test-other}  \\
\midrule
\textbf{(CTC)} Zeghidour et al.\cite{zeghidour2018fullyconv}  & GCNN & 3.3 & 10.5 \\
\textbf{(S2S)}: Mohamed et al. \cite{mohamed2019transformers} & - & 4.7 & 12.9 \\  
\textbf{(S2S)}: Hannun et al. \cite{hannun2019TDS}           & 4-gr & 4.2 & 11.8  \\
\textbf{(S2S)}: Park et al. \cite{park2019specaug}         & RNN & 3.2 & 9.8 \\
~~ + SpecAugment                &  & 2.5 & 5.8 \\
\textbf{(Hybrid)}: Lüscher et al. \cite{luscher2019transformers} & 4-gr & 3.8 & 8.8 \\
~~ + Trf rescoring.                  & Trf  & 2.3 & 5.0 \\
\midrule
\textbf{(CTC)}: Baseline (24 Trf)      &   & 4.0   & 9.4   \\
~~ + Iter. Loss (8-16-24)   & 4-gr   & 3.5   & 8.4    \\
~~ + Feat. Cat. (8-16-24)   &    & 3.3   & 7.6   \\
\cmidrule{2-4}
\textbf{(CTC)}: Baseline (36 Trf)      &   & 4.0   & 9.4   \\
~~ + Iter. Loss (12-24-36)  & 4-gr   & 3.4   & 8.1    \\
~~ + Feat. Cat. (12-24-36)  &   & \textbf{3.2}   & \textbf{7.2}    \\
\midrule
\textbf{(Hybrid)}: Baseline (24 Trf) & & 3.2   & 7.7    \\
~~ + Iter. Loss (8-16-24)   & 4-gr     & 3.1   & 7.3    \\
~~ + Feat. Cat. (8-16-24)   &          & \textbf{2.9}  & \textbf{6.7}   \\
\bottomrule
\end{tabular}
\caption{Librispeech experimental results.  The baseline consists of VGG + 24 layers of Transformers trained with SpecAugment \cite{park2019specaug}. Trf is transformer. 4-gr LM is the official 4-gram word LM. S2S denotes sequence-to-sequence architecture.}
\label{tab:libri-spaug}
\end{center}
\vspace{-0.3cm}
\end{table}

\begin{table}
\begin{center}
\begin{tabular}{lccc}
\toprule
\textbf{Model} & \multicolumn{3}{c}{\textbf{Video}} \\
 & \textbf{curated} & \textbf{clean} & \textbf{noisy} \\
\midrule
\textbf{(CTC)}: Baseline (24 Trf)       & 14.0  & 17.4 & 23.6 \\
~~ + Iter. Loss (8-16-24)               & 13.2  & 16.7 & 22.9 \\
~~ + Feat. Cat. (8-16-24)               & 12.4  & 16.2 & 22.3 \\
\cmidrule{2-4}
\textbf{(CTC)}: Baseline (36 Trf)       & 14.2  & 17.5 & 23.8 \\
~~ + Iter. Loss (12-24-36)              & 12.9  & 16.6 & 22.8 \\
~~ + Feat. Cat. (12-24-36)              & \textbf{12.3}  & \textbf{16.1} & \textbf{22.3} \\
\midrule
\textbf{(Hybrid)}: Baseline (24 Trf)    & 12.8  & 16.1 & 22.1 \\
~~ + Iter. Loss (8-16-24)               & 12.1  & 15.7 & 21.8 \\
~~ + Feat. Cat. (8-16-24)   & \textbf{11.5} & \textbf{15.4} & \textbf{21.4} \\
\bottomrule
\end{tabular}
\caption{Video English dataset experimental results. }
\label{tab:video-spaug}
\end{center}
\vspace{-0.3cm}
\end{table}

As shown in Table \ref{tab:video-spaug}, the proposed methods also provide large performance improvements on the curated video set, up to 13\% with CTC, and up to 9\% with the hybrid model. We also observe moderate gains of between 3.2 and 8\% relative on the clean and noisy video sets.

\section{Related Work}

In recent years, Transformer models have become an active research topic in speech processing. The key features of Transformer networks is self-attention, which produces comparable or better performance to LSTMs when used for encoder-decoder based ASR \cite{sperber2018self}, as well as when trained with CTC \cite{karita2019comparative}. Speech-Transformers \cite{dong2018speech} also produce comparable performance to the LSTM-based attention model, but with higher training speed in a single GPU. Abdelrahman et al.\cite{mohamed2019transformers} integrates a convolution layer to capture audio context and reduces WER in Librispeech. 

The use of an objective function in intermediate layers has been found useful in several previous works such as image classification \cite{szegedy2015going} and language modeling \cite{al2019character}. In \cite{Rao_2017}, the authors did pre-training with an RNN-T based model by using a hierarchical CTC criterion with different target units. In this paper, we don't need additional types of target unit, instead we just use same tokenization and targets for both intermediate and final losses.

The application of the objective function to intermediate layers is also similar in spirit to the use of KL-divergence in \cite{lu2019self}, which estimates output posteriors at an intermediate layer and regularizes them towards the distributions at the final layer. In contrast to this approach, the direct application of the objective function does not require the network to have a good output distribution before the new gradient contribution is meaningful.

\section{Conclusion}
In this paper, we have proposed a method for re-processing the input features in light of the information available at an intermediate network layer. We do this in the context of deep transformer networks, via a self-attention mechanism on both features and hidden states representation. To encourage meaningful partial results, we calculate the objective function at intermediate layers of the network as well as the output layer. This improves performance in and of itself, and when combined with feature re-presentation we observe consistent relative improvements of 10 - 20\% for Librispeech and 3.2 - 13\% for videos.

\vfill\pagebreak

\bibliographystyle{IEEEbib}
\bibliography{DEEPTRF_v1}

\end{document}